\documentclass[10pt, a4paper]{article}

\usepackage{lrec-coling2024} % this is the new style
\usepackage{subcaption}
\usepackage{amsmath, amssymb, amsthm}
\usepackage{xcolor}
\definecolor{myred}{RGB}{248,212,214}
\definecolor{myblue}{RGB}{187,215,232}
\definecolor{mypurple}{RGB}{230,216,234}
\definecolor{myyellow}{RGB}{251,230,191}
\definecolor{mygreen}{RGB}{233,242,219}

\title{CBT-LLM: A Chinese Large Language Model for Cognitive Behavioral Therapy-based Mental Health Question Answering}

\name{Hongbin Na} 

\address{Australian Artificial Intelligence Institute, \\University of Technology Sydney, Sydney, Australia \\
        hongbin.na@student.uts.edu.au
         }
         
\abstract{
The recent advancements in artificial intelligence highlight the potential of language models in psychological health support. While models trained on data from mental health service platform have achieved preliminary success, challenges persist in areas such as data scarcity, quality, and ensuring a solid foundation in psychological techniques. To address these challenges, this study introduces a novel approach to enhance the precision and efficacy of psychological support through large language models. Specifically, we design a specific prompt derived from principles of Cognitive Behavioral Therapy (CBT) and have generated the CBT QA dataset, specifically for Chinese psychological health Q\&A based on CBT structured intervention strategies. Unlike previous methods, our dataset emphasizes professional and structured response. Utilizing this dataset, we fine-tuned the large language model, giving birth to CBT-LLM, the large-scale language model specifically designed for Cognitive Behavioral Therapy techniques. Empirical evaluations demonstrate that CBT-LLM excels in generating structured, professional, and highly relevant responses in psychological health support tasks, showcasing its practicality and quality. The model is available on Hugging Face: \url{https://huggingface.co/Hongbin37/CBT-LLM}.
 \\ \newline  }

\begin{document}

\maketitleabstract

\section{Introduction}
\noindent
\textbf{\textcolor{red}{Important:}} \textcolor{red}{Our research explores the potential of large language models to answer questions based on Cognitive Behavioral Therapy, but does \textbf{NOT} recommend their use as a substitute for psychological treatment without professional supervision.}
% In recent years, there has been a growing recognition of the remarkable capabilities of pre-trained language models in various domains \cite{shah-etal-2022-flue, Madani2023LargeLM}. One area that has received significant attention is the intersection of artificial intelligence and mental health. Particularly, to utilize pre-trained language models to offer mental health support for users, multiple systems for mental health based on pre-trained language models have been proposed \cite{liu-etal-2021-towards, cheng-etal-2023-pal}, and achieved preliminary results. The efficacy of these models within the context of mental health can be primarily attributed to their deep architectures and attention mechanisms, enabling them to capture nuanced information related to emotions and cognition. To date, pre-trained language models have demonstrated significant potential in reading, processing, and understanding natural language texts associated with mental health. 

The advancement of pre-trained language models (PLMs) has profoundly impacted various domains, such as finance \cite{shah-etal-2022-flue}, biology \cite{Madani2023LargeLM}, heralding new frontiers in applications and research. Among these, the intersection of artificial intelligence and mental health has emerged as a particularly promising area. Here, the deployment of PLMs holds the potential to revolutionize mental health support, a notion underscored by the development of several preliminary systems aimed at leveraging these models for psychological assistance \cite{liu-etal-2021-towards, cheng-etal-2023-pal, Lai2023PsyLLMSU}. The core strengths of PLMs, namely their deep learning architectures and sophisticated attention mechanisms, enable them to parse and interpret complex emotional and cognitive information, positioning them as invaluable tools in mental health contexts.

% 现有方法
% Utilizing pre-trained language models for mental health support is challenging, primarily due to the scarcity of data. PsyQA \cite{sun-etal-2021-psyqa} is a dataset that addresses the lack of Chinese mental health data by providing a structured question-answer pair dataset scraped from Chinese mental health service platforms. To tackle the difficulty of collecting large-scale and genuine multi-turn mental health conversations, \citet{Qiu2023SMILEST} proposed SMILE, a strategy that leverages ChatGPT to expand public single-turn conversations into multi-turn dialogues. Additionally, to address challenges posed by more authentic counseling scenarios, Psy-LLM \cite{Lai2023PsyLLMSU} incorporated numerous psychology articles into its training. However, the above methods are still far from providing sufficiently precise mental health support.

Despite these advancements, the application of PLMs in mental health support is fraught with challenges, particularly regarding data scarcity and quality. The PsyQA dataset \cite{sun-etal-2021-psyqa} attempts to mitigate the dearth of Chinese mental health data by collating structured question-answer pairs from online services. Similarly, the SMILE strategy \cite{Qiu2023SMILEST} employs ChatGPT to transform single-turn conversations into multi-turn dialogues, addressing the shortage of authentic multi-turn mental health discussions. Further, Psy-LLM \cite{Lai2023PsyLLMSU} enriches its dataset with psychology articles to mimic more realistic counseling scenarios. Nevertheless, these approaches still fall short in delivering precise and effective mental health support.

\begin{figure}[t]
    \centering
    \includegraphics[width=1\linewidth]{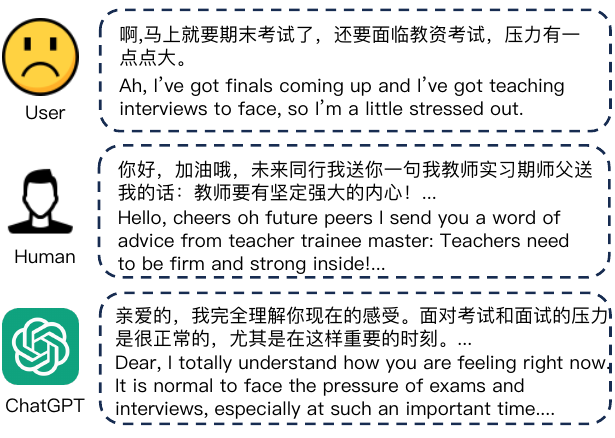}
    \caption{An example of poor data quality. "Human" represents actual human responses from an online mental health forum, while "ChatGPT" denotes our responses generated based on Cognitive Behavioral Therapy (CBT) prompt.}
    \label{fig:data}
\end{figure}

% 研究挑战
% Delving deeper into the challenges, in the domain of mental health support, the design and implementation of Q\&A system face a series of complications. Firstly, for the construction of effective mental health Q\&A systems, merely relying on a large volume of data is insufficient; the quality of the data is equally crucial. Taking PsyQA as an example, it depends on responses from a mental health service platform as its prediction target. However, these responses do not always play a positive role in assisting users. As shown in Fig. \ref{fig:data}, when users seek help due to exam stress, some responses might lack empathy and understanding, sometimes even adopting a critical attitude towards the user's situation. Secondly, although the objective of mental health dialogue systems is to provide effective support for users, current systems \cite{Lai2023PsyLLMSU, Qiu2023SMILEST} often aren't fully grounded in professional psychological principles and techniques. This detachment could result in limitations when offering support. More importantly, to enable models to provide responses based on psychological techniques, data specifically based on psychological counseling techniques is required. However, such data is still scarce at present.

A critical examination of existing mental health Q\&A systems reveals inherent complexities in their development and deployment. For instance, the reliance on large datasets does not necessarily translate to high-quality mental health support, as exemplified by the PsyQA dataset \cite{sun-etal-2021-psyqa}, where responses from professional platforms may not always be constructive or empathetic, particularly in stress-related scenarios (see Fig. \ref{fig:data}). Moreover, the absence of grounding in established psychological methodologies in current dialogue systems leads to a gap between the support provided and professional mental health standards. This gap underscores the pressing need for data rooted in bona fide psychological counseling techniques, which remains scarce.

% 创新思路
In response to these challenges, this study introduces a novel approach dedicated to enhancing the accuracy and effectiveness of psychological health support. Initially, we designed a specific prompt for Cognitive Behavioral Therapy (CBT) counseling techniques based on the principles of CBT. As a validated and effective psychological treatment method \cite{Hofmann2012TheEO, David2018WhyCB}, the structured intervention strategies of CBT provided us with a theoretical foundation to build efficient dialogue models. Leveraging this prompt, we further developed the CBT QA dataset. This dataset is specially designed for chinese mental health dialogues, aiming to provide questions and answers based on the structure of CBT. Unlike previous methods that relied on public platform data, our dataset is more professional and structured, ensuring that model responses align more closely with psychological principles and practices. Most critically, based on the CBT QA dataset, we fine-tuned the large language model (LLM) with instructions, successfully establishing CBT-LLM, specifically designed for cognitive behavioral therapy techniques.

% 实验结论
% After multiple rounds of experiments and evaluations, our CBT-LLM model demonstrated significant performance in tasks related to psychological health support. Through automatic evaluation metrics, our model distinctly exhibited its strict adherence to the structure of CBT responses, ensuring that the generated answers are not only structured but also professional. Additionally, in manual evaluations, the answers produced by our model were highly relevant to the questions posed by users, offering them substantial assistance and information. This further confirms the notable effectiveness of our approach in ensuring the quality and practicality of responses.

Our comprehensive experiments and evaluations demonstrate that the proposed CBT-LLM model significantly enhances support tasks in tasks related to psychological health support. It not only adheres strictly to CBT structural guidelines but also delivers responses that are professional, structured, and highly relevant to users' needs. These findings, substantiated by both automatic and manual assessments, highlight the efficacy and practicality of our approach.

% The main contributions are summarized as follows:
% \begin{itemize}
%     \item In this study, we designed a specific prompt based on the principles of CBT and developed the CBT QA dataset, tailored for Chinese mental health single-turn Q\&A.
%     \item Building upon the CBT QA dataset, we fine-tuned the LLM, leading to the successful establishment of CBT-LLM, the first large language model specifically designed for cognitive behavioral therapy techniques.
%     \item Experiments show that our CBT-LLM performs excellently in mental health support tasks. Whether through automatic evaluation or manual assessment, the model's answers are highly relevant to user questions, offering significant help and information.
% \end{itemize}
The contributions of this study are threefold:
\begin{itemize}
    \item The design of a novel CBT-based prompt and the development of the CBT QA dataset, specifically tailored for Chinese mental health dialogues.
    \item The adaptation of a large language model into CBT-LLM, leveraging the CBT QA dataset for nuanced mental health support, marking a pioneering step in applying PLMs to cognitive behavioral therapy.
    \item Comprehensive validation showing that CBT-LLM significantly outperforms existing models in providing mental health support, as evidenced by both automatic metrics and human evaluation.
\end{itemize}

\section{Related Work}
\subsection{Counseling Techniques}
Psychological counseling techniques offer significant support for individual mental health and quality of life, facilitating individuals in identifying, resolving, and coping with psychological issues and dilemmas \cite{meier2010counselling}. Cognitive Behavioral Therapy (CBT), one of the most widely used approaches \cite{beck1979cognitive}, focuses on the interplay of cognition, emotion, and behavior in influencing mental health. Therapists work with clients to identify and challenge harmful thought patterns, also known as cognitive distortions, promoting healthier coping mechanisms. Acceptance and Commitment Therapy (ACT), representing the third wave of cognitive behavioral therapies \cite{hayes2003acceptance}, emphasizes psychological flexibility and embracing present experiences while acting in line with personal values. On the other hand, humanistic psychotherapy, such as Carl R. Rogers' client-centered therapy \cite{rogers1951client}, centers on individuals' self-actualization and growth, with therapists providing unconditional positive regard, empathy and genuineness to support self-discovery and personal development. Dialectical Behavior Therapy (DBT), developed by \citet{linehan2014dbt} as a modification of CBT for treating borderline personality disorder, integrates cognitive-behavioral techniques with mindfulness practices from Buddhist traditions, aiming to balance acceptance and change to improve emotional regulation and interpersonal effectiveness.

\subsection{LLMs for Mental Health Support}
LLMs have gained significant attention across various research domains, including medicine \cite{Thirunavukarasu2023LargeLM}, education \cite{Dan2023EduChatAL}, and finance \cite{Wu2023BloombergGPTAL}. In the field of mental health support, the use of LLMs is an emerging and valuable research area \cite{Dhingra2023MindMM}. Recent studies have shown that ChatGPT, in particular, excels in mental health analysis and model interpretability when compared to traditional neural network approaches \cite{Yang2023TowardsIM}. To address the limited availability of mental health data, researchers have created the ExTES emotion support dialogue dataset, while the SMILE approach extends single-turn dialogues to multi-turn interactions, enriching the data source for mental health support \cite{Zheng2023BuildingES, Qiu2023SMILEST}. Additionally, a framework called Psy-LLM has been proposed to provide real-time feedback to mental health professionals by combining pre-trained LLMs with psychological forum Q\&A \cite{Lai2023PsyLLMSU}. Furthermore, the rapid development of LLM has promoted psychological counseling, which mainly includes motivational interviewing \cite{min-etal-2022-pair, welivita-pu-2023-boosting} and cognitive behavioral therapy \cite{ding-etal-2022-improving, maddela-etal-2023-training, sharma-etal-2023-cognitive}, where LLMs offer new methodologies for delivering interventions and support. 

\section{Methodology}
% 画一个方法的Overview
\begin{figure}
    \centering
    \includegraphics[width=1\linewidth]{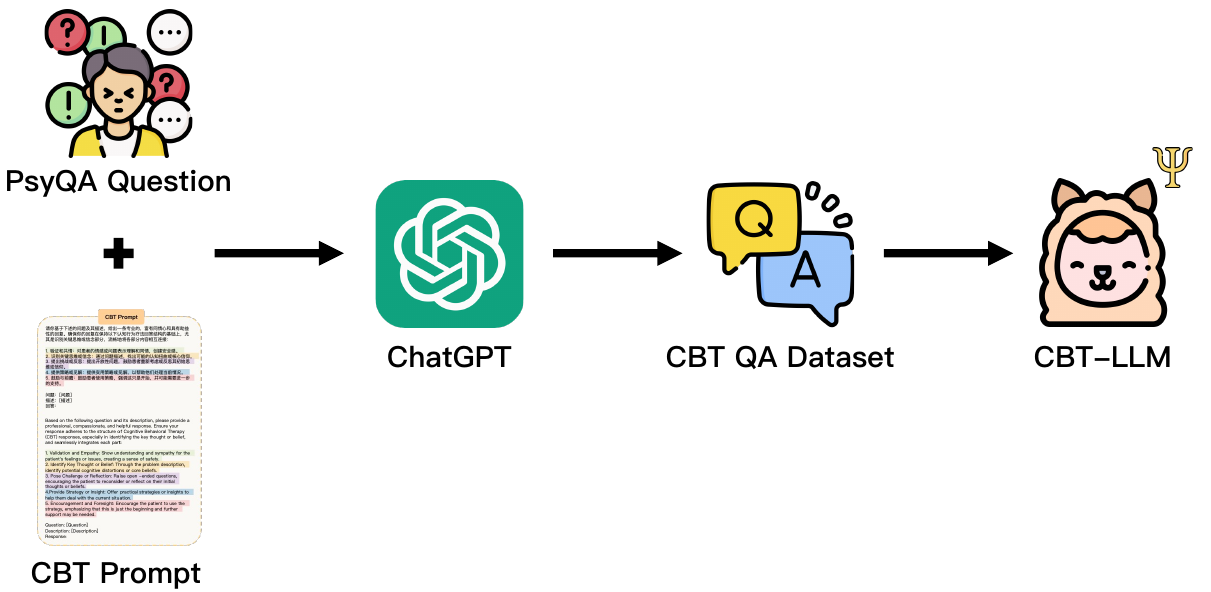}
    \caption{An overview of training CBT-LLM. It first utilizes PsyQA Questions and CBT Prompt to generate CBT answers, and then fine-tuning CBT-LLM.}
    \label{fig:overview}
\end{figure}

\subsection{Problem Definition}
Given a cognitive psychology question-answer dataset \textit{PsyQA} \cite{sun-etal-2021-psyqa} with questions and their descriptions and a set of CBT prompts represented by \( \mathcal{P}_{\text{CBT}} \), we employ \textit{ChatGPT} in conjunction with \( \mathcal{P}_{\text{CBT}} \) to generate CBT-oriented answers. For each question \( q_i \) and its description \( d_i \) from \textit{PsyQA}, the CBT response is derived as \( a_i = \text{ChatGPT}(q_i, d_i, \mathcal{P}_{\text{CBT}}) \). Assembling the questions, descriptions, and generated answers results in the CBT QA dataset, represented as \( \mathcal{D}_{\text{CBT QA}} = \{(q_i, d_i, a_i)\} \). This dataset subsequently undergoes instruction fine-tuning to cultivate the specialized CBT-LLM. The overarching aim is to utilize the insights from the \textit{PsyQA} dataset in tandem with \( \mathcal{P}_{\text{CBT}} \) through the intervention of \textit{ChatGPT} to derive a language model proficient in CBT question-answering.

\subsection{Generation of CBT Responses}
% CBT has been widely recognised as an effective psychological treatment
CBT is a well-established psychological intervention
for a wide range of psychological disorders. Despite the importance of CBT in mental health practice, current mental health support datasets for the development of language models do not yet adequately cover this area. To address this gap, we turned to PsyQA \cite{sun-etal-2021-psyqa}, a well-known mental health Q\&A dataset. The dataset is derived from the Chinese online mental health support forum Yixinli\footnote{\url{https://www.xinli001.com/qa}}. The dataset encapsulates a broad spectrum of pertinent and complex questions.
% and covers a wide range of representative and challenging questions. 
Given the extensive variety of queries in \textit{PsyQA}, crafting individual professional CBT responses is not feasible. We draw inspiration from the Alpaca study \cite{alpaca}, which demonstrated the high quality of data generated by ChatGPT. Building upon this precedent, we meticulously designed CBT-centric prompts to guide ChatGPT in providing CBT-informed responses to the questions in \textit{PsyQA}. To ensure robust generative capabilities and manage context length effectively, we utilized OpenAI's gpt-3.5-turbo-16k model\footnote{\url{https://openai.com/blog/chatgpt}}.

% Given the vast diversity of questions in PsyQA, individual professional CBT responses are impractical. Inspiration is drawn from the Alpaca \cite{alpaca} study. Alpaca demonstrated that data generated by ChatGPT possess high quality by leveraging it to produce the required data. Building on this approach, we meticulously crafted prompts grounded in CBT principles to utilize ChatGPT in providing CBT-based answers for questions in PsyQA. 

\begin{figure}
    \centering
    \includegraphics[width=1\linewidth]{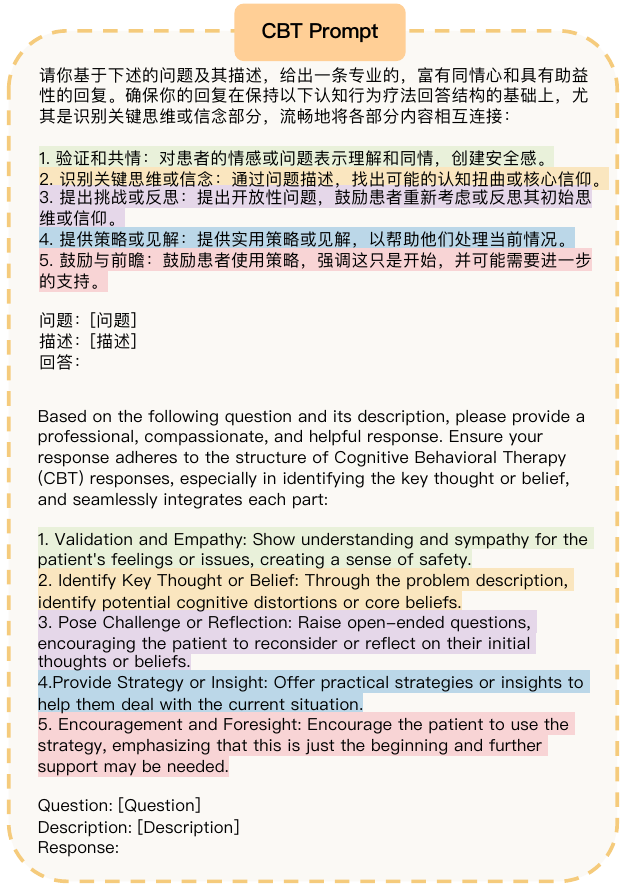}
    \caption{CBT prompt for dataset generation.}
    \label{fig:cbt}
\end{figure}
The foundational aspects of CBT are outlined in \citet{beck1979cognitive} research. The primary objective of CBT is to identify and comprehend an individual's automatic thoughts and core beliefs, which play a crucial role in shaping their emotional and behavioral disturbances. CBT places great emphasis on challenging distorted cognitions, aiming to rectify any cognitive biases that contribute to psychological distress. To facilitate lasting change, CBT incorporates behavioral experiments and skill training, encouraging individuals to implement and practice new strategies in real-world settings. 
Recognizing the essential elements of CBT, it is crucial to adopt a comprehensive approach when applying these methodologies and strategies in therapeutic contexts.
% Given the foundational components of CBT, it becomes imperative to adopt an integrative approach when employing these methodologies and strategies in therapeutic interventions.

% Drawing upon the core tenets of CBT, we have restructured and formulated the response mechanism into the following five key points, tailored to better accommodate the single-turn dialogue response paradigm:
In alignment with the core principles of CBT, we have restructured and formulated the response mechanism into five pivotal components, specifically adapted to suit the single-turn dialogue response format:
\begin{enumerate}
    \item \textbf{Validation and Empathy}: Show understanding and sympathy for the patient's feelings or issues, creating a sense of safety.
    \item \textbf{Identify Key Thought or Belief}: Through the problem description, identify potential cognitive distortions or core beliefs.
    \item \textbf{Pose Challenge or Reflection}: Raise open-ended questions, encouraging the patient to reconsider or reflect on their initial thoughts or beliefs.
    \item \textbf{Provide Strategy or Insight}: Offer practical strategies or insights to help them deal with the current situation.
    \item \textbf{Encouragement and Foresight}: Motivate the individual to employ the suggested strategy, underscoring that this is merely an initial step and additional support may be warranted.
    % Encourage the patient to use the strategy, emphasizing that this is just the beginning and \bei{additional} support may be needed.
\end{enumerate}

Based on the aforementioned structure, we crafted the prompt depicted in Fig. \ref{fig:cbt}, aiming to steer ChatGPT towards generating responses that are congruent with CBT methodologies.
% with the intention of guiding ChatGPT to provide responses in accordance with the principles and framework of CBT. 
% To ensure the stability of prompt outputs, we randomly selected a question and its description and passed it to GPT-4 through the CBT Prompt, obtaining a sample CBT response. Subsequently, we incorporated this example into the prompt for generating subsequent replies. This approach aims to ensure that our prompts effectively guide ChatGPT in producing answers that align with CBT principles.
To verify the consistency of the prompt outputs, we randomly selected a question with its description and submitted it to GPT-4 via the CBT Prompt, procuring a prototype CBT response. This example was then incorporated into the prompt to inform the generation of subsequent responses. This iterative refinement ensures that our prompts effectively direct ChatGPT to produce answers that are consistent with CBT principles.

\subsection{CBT Response Analysis}
\subsubsection{General Statistical Analysis}
Our dataset, detailed in Table \ref{table:dataSA}, consists of 22,327 entries, each comprising questions, descriptions, and CBT responses. On average, questions are concise, with 21.6 characters, whereas descriptions are more extensive, averaging 168.9 characters. The CBT responses, with an average of 522.8 characters, demonstrate the need for more elaborate text to provide effective advice and explanations. Our analysis further delves into the prevalence of cognitive dissonance within these responses, identifying its presence in 12,136 instances, or 54.4\% of the dataset. This emphasises the widespread presence of cognitive dissonance in counseling scenarios.
% As shown in Table \ref{table:dataSA}, the dataset comprises 22,327 questions and descriptions, with a corresponding number of 22,327 CBT responses. On average, each question contains 21.6 characters, while descriptions tend to be longer, with an average of 168.9 characters. The average character count for CBT responses is 522.8, indicating that CBT responses typically require more text for explanation and advice. Furthermore, we analyzed the presence of cognitive dissonance in CBT responses. The results showed that 12,136 samples exhibited cognitive dissonance, representing approximately 54.4\% of the total text data. This emphasises the widespread presence of cognitive dissonance in counseling scenarios.

\begin{table}[t]
\renewcommand{\arraystretch}{1.2}
\begin{tabular}{cc}
\hline
\textbf{Criteria}              & \textbf{Statistics} \\ \hline
No. of Question \& Description & 22327               \\
No. of CBT Response            & 22327               \\ \hline
Characters Per Question        & 21.6                \\
Characters Per Description     & 168.9               \\
Characters Per CBT Response    & 522.8               \\ \hline
Percentage of Cognitive Distortions & 54.4\%          \\ \hline
\end{tabular}
\caption{CBT QA dataset statistical analysis.}
\label{table:dataSA}
\end{table}

\begin{table*}[t]
\resizebox{\textwidth}{!}{
\renewcommand{\arraystretch}{1.2}
\begin{tabular}{ccc}
\hline
\textbf{Cognitive Distortion Type} & \textbf{Interpretation}                                                                                                                                                                                                                               & \textbf{Samples} \\ \hline
All-or-Nothing Thinking            & \begin{tabular}[c]{@{}c@{}}The tendency to see things as extremes, \\ either total success or total failure, \\ ignoring the possibilities in between.\end{tabular}                                                                                   & 7115             \\ \hline
Overgeneralization                 & \begin{tabular}[c]{@{}c@{}}It makes unreasonable inferences about the general \\ situation based on limited experience or a single event, \\ usually based on negative experience.\end{tabular}                                                       & 7782             \\ \hline
Emotional Reasoning                & \begin{tabular}[c]{@{}c@{}}It manifests itself by judging things based on \\ one's emotional state without relying on\\ objective evidence or logical reasoning.\end{tabular}                                                                         & 742              \\ \hline
Catastrophizing                    & \begin{tabular}[c]{@{}c@{}}It involves exaggerating the seriousness of errors \\ or problems, seeing them as great disasters, \\ and usually ignoring the reasonableness of the actual situation.\end{tabular}                                        & 349              \\ \hline
Mind Reading                       & \begin{tabular}[c]{@{}c@{}}It manifests itself in the false assumption that one knows what others are \\ thinking or feeling, without relying on clear communication or evidence, \\ and usually leads to misunderstanding and conflict.\end{tabular} & 345              \\ \hline
Others                             & \begin{tabular}[c]{@{}c@{}}Contains Fortune Telling, Filtering, Attribution Error, \\ and other types of cognitive distortions.\end{tabular}                                                                                                          & 2094             \\ \hline
\end{tabular}}
\caption{Cognitive distortion statistical analysis on CBT QA dataset.}
\label{table:CDSA}
\end{table*}

\subsubsection{Cognitive Distortion Statistical Analysis}
In Table \ref{table:CDSA}, we present a comprehensive statistical analysis of the 12,136 instances that display cognitive distortions. Drawing from \citet{beck2020cognitive}, we categorize and understand the nature of these distortions.
% Table \ref{table:CDSA} presents a detailed statistical analysis of 12,136 samples exhibiting cognitive distortions. These samples represent various types of cognitive distortions that individuals exhibit in their thinking and emotions. We extracted explanations for each type of cognitive distortion from the work of \citet{beck2020cognitive} to better understand them. 
Firstly, we observed that All-or-Nothing Thinking is among the most prevalent types, accounting for approximately 59\% of the total samples, encompassing 7,115 samples. This distortion is characterized by a tendency to see things in black and white, viewing them as either complete successes or utter failures, neglecting the existence of intermediary possibilities. Secondly, Overgeneralization represents about 64\% of the total samples, totaling 7,782 samples. This distortion involves making unwarranted inferences about the bigger picture based on limited experiences or a single event, often stemming from negative experiences. Moreover, we noticed that individual samples can exhibit multiple cognitive distortions, with All-or-Nothing Thinking often co-occurring with Overgeneralization. Lastly, Other significant distortions, such as Emotional Reasoning, Catastrophizing, and Mind Reading, account for 29\% of the cases, showcasing a diverse range of erroneous thinking patterns in the sample.
% other cognitive distortion types, including Emotional Reasoning, Catastrophizing, and Mind Reading, which don't encompass the aforementioned two, represent 29\% of the total samples.

\subsubsection{Recognition of Quality Analysis}
% To evaluate the performance of CBT Prompt in recognizing cognitive distortions, we employed three evaluation metrics: accuracy, recall, and F1 score. Given that the original dataset did not have genuine labels regarding cognitive distortions, we randomly selected 500 samples from the generated ones. Subsequently, professional psychotherapists were invited to manually annotate them to determine the presence of cognitive distortions in each sample. Detailed evaluation results are presented in Table \ref{table:RQA}. From the results, it can be observed that the accuracy of CBT Prompt reached 0.69, which is relatively high, indicating its good prediction correctness. Moreover, with a recall of 0.93, it suggests that our approach can successfully detect the vast majority of real instances of cognitive distortions. However, an accuracy of 0.69 and an F1 Score of 0.65 also hint at a certain level of false positives in the predictions.

We evaluated the effectiveness of the CBT Prompt in identifying cognitive distortions using accuracy, recall, and F1 score as metrics. Owing to the lack of pre-labeled data for cognitive distortions, we curated a subset of 500 randomly selected samples. These were then annotated by professional psychotherapists to serve as a ground truth for performance assessment, with the detailed outcomes presented in Table \ref{table:RQA}. The CBT Prompt demonstrated commendable accuracy (0.69), suggesting a high level of correctness in predictions. The recall rate of 0.93 indicates the system's proficiency in identifying the majority of actual cognitive distortion instances. However, the juxtaposition of a 0.69 accuracy rate and a 0.65 F1 Score indicates the presence of false positives, pointing to areas for improvement in our prediction model.

\begin{table}[t]
\centering
\renewcommand{\arraystretch}{1.2}
\begin{tabular}{cccc}
\hline
                    & \textbf{Accuracy} & \textbf{Recall} & \textbf{F1 score} \\ \hline
Quality & 0.69     & 0.93   & 0.65              \\ \hline
\end{tabular}
\caption{Cognitive distortion of recognition quality.}
\label{table:RQA}
\end{table}

\subsection{CBT-LLM}
In our research on automating Cognitive Behavioral Therapy (CBT) question-answer tasks, we chose to utilize Language Models (LMs) as the foundational framework, specifically leveraging the power of Large Language Models (LLMs) due to their exceptional performance across various Natural Language Processing (NLP) tasks. However, to better tailor the LLMs to the specific requirements of CBT Q\&A, we incorporated advanced fine-tuning strategies, instruction tuning \cite{Wang2022SelfInstructAL} and LoRA \cite{Hu2021LoRALA}.

% At the core of LLMs of Transformer-Decoder only lies an autoregressive structure, where the model aims to predict the next word in a given sequence. This process can be represented by the following equation:
The fundamental architecture of the employed LLMs is based on a Transformer-Decoder, characterized by an autoregressive framework that sequentially predicts each subsequent word. The mathematical representation of this process is articulated as:
\begin{equation}
P(x_1, x_2, \dots, x_T) = \prod_{t=1}^{T} P(x_t | x_1, \dots, x_{t-1})
\end{equation}
\noindent This formulation elucidates that the likelihood of a word sequence is determined by multiplying the conditional probabilities of each subsequent word, predicated on all preceding ones.
% This equation demonstrates that the probability of a sequence of words is calculated as the product of the conditional probabilities of each word given the preceding words.

To fully exploit the potential of LLMs for complex tasks like CBT Q\&A, we incorporated instruction tuning and LoRA fine-tuning strategy. Instruction tuning is a method that provides explicit task directives to the model during training, guiding its generation process to align more closely with the specific task requirements. We adopted the approach proposed by \citet{Wang2022SelfInstructAL} for instruction tuning. Moreover, LoRA is a fine-tuning strategy that enhances model performance by augmenting each layer of the model with additional parameters. These parameters are represented using low-rank matrices, effectively adjusting the output of each layer. Specifically, for a pre-trained weight matrix \( W \), its parameter update \( \Delta W \) can be calculated as:
\begin{equation}
\Delta W = A \times B
\end{equation}
Here, \( A \) and \( B \) are low-rank matrices, while the original parameter \( W \) remains unchanged throughout the training process.

In addition, for training the CBT-LLM, we used cross-entropy loss, a commonly used loss function for language modeling tasks. The cross-entropy loss measures the dissimilarity between the predicted probability distribution and the true distribution of the next word. The loss can be computed using the following equation:
\begin{equation}
\text{Loss} = -\sum_{t=1}^{T} \log P(x_t | x_1, \dots, x_{t-1})
\end{equation}
In summary, by incorporating instruction tuning and LoRA fine-tuning strategy, we enhance the adaptability and performance of the LLMs to better meet the requirements of CBT Q\&A tasks. %The autoregressive nature of LLMs, combined with these advanced techniques, enables the model to generate more contextually relevant and accurate responses.

\section{Experiment}

\subsection{Data Preparation}
We selected the CBT QA dataset and randomly split it into a training set (90\%) and a test set (10\%). To meet the data format requirements for instruction-based fine-tuning, we concatenated the question and description to form a single passage as input, using the CBT response as output. We further incorporated the instruction: "You are an experienced therapist specializing in cognitive behavioral therapy. Please answer the following questions in the capacity of a psychotherapist," forming a triad format of \textit{\{"instruction": , "input": , "output"\}}.

\subsection{Baselines}
\begin{itemize}
\item\textbf{LLaMA-Chinese-7B:} The LLaMA-Chinese-7B \cite{chinese-llama-alpaca} is derived from the LLaMA-7B \cite{Touvron2023LLaMAOA} model, enhanced with a series of optimizations specific to Chinese processing. Researchers expanded the model's vocabulary, adding 20,000 Chinese tokens, culminating in a Chinese LLaMA tokenizer with a vocabulary size of 49,953. Subsequently, the model underwent secondary pre-training on 20GB of Chinese data, and by adopting the LoRA approach, freezing weights and incorporating low-rank matrices, the training efficiency was improved.

\item\textbf{Alpaca-Chinese-7B:} 
Building on the foundation laid by LLaMA-Chinese-7B, the Alpaca-Chinese-7B model \cite{chinese-llama-alpaca} underwent further refinement to specialize in instruction-following capabilities. This development utilized an instruction dataset ranging from 2M to 3M entries, aiming to improve the model's performance in executing user-specified tasks.
% Building upon the successful construction of the LLaMA-Chinese-7B model, researchers further fine-tuned it with the aim of creating a model that follows instructions, termed Alpaca-Chinese-7B \cite{chinese-llama-alpaca}. This refinement process utilized approximately 2M to 3M instruction data.

\item\textbf{Qwen-7B:}
With a training corpus exceeding 2.4 trillion Chinese tokens, Qwen-7B \cite{Bai2023QwenTR} encompasses English and other languages. The model exhibits outstanding performance in a variety of downstream tasks. Qwen-7B utilizes a vocabulary size of approximately 150,000 tokens. Compared to mainstream open-source Chinese-English vocabulary models, it is better suited for multilingual processing, allowing users to enhance and extend the processing capability for specific languages without the need to expand the vocabulary.

\item\textbf{Baichuan-7B:}
Engineered with the Transformer architecture, Baichuan-7B \cite{Baichuan7B2023} leverages 7 billion parameters and was trained on a 1.2 trillion token corpus. It uniquely supports extended text sequences with a context window of 4096 and excels in Chinese language processing while maintaining effective bilingual (Chinese and English) support. Baichuan-7B has shown exemplary results on the C-EVAL benchmark \cite{huang2023ceval}, making it a highly suitable model for CBT Q\&A applications, particularly for tasks requiring extensive comprehension and generation in Chinese.
On the authoritative Chinese benchmark C-EVAL \cite{huang2023ceval}, Baichuan-7B has demonstrated superior performance in Chinese language tasks. With its unique design characteristics, Baichuan-7B is considered an ideal choice for CBT Q\&A implementations.

% Baichuan-7B \cite{Baichuan7B2023} is built on the Transformer architecture, boasting 7 billion parameters, and was trained on approximately 1.2 trillion tokens. Notably, this model supports both Chinese and English languages and has a context window length of 4096, allowing it to handle extended texts. Unique features of Baichuan-7B particularly excel in Chinese language processing. While it offers bilingual support, it is specially optimized for Chinese, indicating an enhanced capability in understanding and generating content in Chinese. 
% Additionally, the model incorporates rotary-embedding for position encoding, as detailed in \cite{su2021roformer}, a technique widely adopted in contemporary models due to its exceptional extrapolation capability. 

% 根据baichuan的训练方式再补充一些内容
\end{itemize}

\begin{table*}[t]
\renewcommand{\arraystretch}{1.2}
\centering
\begin{tabular}{c|ccccc}
\hline
\textbf{CBT-LLM Backbone}    & \textbf{BLEU}   & \textbf{METEOR} & \textbf{CHRF}   & \textbf{BLEURT} & \textbf{BERTSCORE} \\ \hline
LLama-Chinese-7B  & 0.2412          & 0.3758          & 0.2167          & 0.5091          & 0.7793             \\
Alpaca-Chinese-7B & 0.2607          & 0.3991          & 0.2596          & 0.5216          & \textbf{0.7849}    \\
Qwen-7B           & 0.2361          & 0.3726          & 0.2939          & 0.5096          & 0.7802             \\ \hline
\textbf{Baichuan-7B}  & \textbf{0.2648} & \textbf{0.4031} & \textbf{0.3839} & \textbf{0.5247} & 0.7841             \\ \hline
\end{tabular}
\caption{Automatic evaluation results on CBT QA dataset.}
\label{table:aer}
\end{table*}

\subsection{Experimental Setups}
In this experiment, we utilized the NVIDIA V100 32G GPU for model training. During training, we set the gradient accumulation steps to 4, meaning that the gradients from every 4 batches would be accumulated and then used for a single parameter update. The learning rate was set to $5 \times 10^{-5}$, and we adopted the cosine-type learning rate scheduler to adjust the learning rate throughout the training process. The entire training will span across 3 epochs. To accelerate the training and enhance model performance, we also enabled the use of 16-bit half-precision floating point numbers. It's noteworthy to mention that the fine-tuning implementation for this model is based on LLaMA Factory \cite{llama-factory}, an efficient model tuning toolset.

\subsection{Automatic Evaluation}
To comprehensively evaluate the model's performance on mental health support Q\&A tasks, we adopted a series of automatic evaluation metrics. BLEU \cite{Papineni2002BleuAM} mainly assesses the precise matching between model outputs and reference answers by comparing n-gram co-occurrences. METEOR \cite{Banerjee2005METEORAA} goes beyond precise matching, incorporating matches at the level of synonyms, stems, and morphological variations, offering a more holistic assessment of semantic similarity. CHRF \cite{popovic-2015-chrf}, a character-based metric, primarily gauges the alignment at the character level between model outputs and reference answers. Both BLEURT \cite{sellam-etal-2020-bleurt} and BERTSCORE \cite{bert-score} utilize pre-trained representations from the BERT \cite{devlin-etal-2019-bert} model to evaluate deep semantic alignment. While BLEURT is tailored for assessing outputs from machine translation and text generation tasks, BERTSCORE calculates the cosine similarity between BERT embeddings of the model output and references. To ensure the accuracy and consistency of our evaluation, we employed the July toolkit \cite{cavusoglu2023jury} for the computation of these automatic evaluation metrics. The results of the automatic evaluation are depicted in Table \ref{table:aer}.

\subsection{Main Results}
Our experiments demonstrate the superiority of our CBT-LLM model in the domain of mental health support Q\&A tasks, outperforming three advanced benchmark models. 
% Based on the experimental results, in the mental health support Q\&A task, our CBT-LLM model demonstrated superior performance compared to the other three advanced models.
Firstly, LLaMA-Chinese-7B, as a Chinese-optimized version of LLaMA-7B, underwent various optimizations for Chinese processing. However, its performance on tasks related to CBT structured interventions still lagged behind CBT-LLM. This can be attributed to the fact that its training and optimization strategies were not specifically tailored for mental health support tasks.
In contrast, Alpaca-Chinese-7B, which builds on the foundation of LLaMA-Chinese-7B, shows improved performance due to its fine-tuning geared towards adherence to CBT principles. Nevertheless, it still slightly lagged behind our CBT-LLM, underscoring the tailored model’s specialized effectiveness.
% In contrast, Alpaca-Chinese-7B, built upon LLaMA-Chinese-7B, underwent further fine-tuning, especially in terms of guidance. Such fine-tuning might have enabled it to better understand and adhere to CBT's structured intervention strategies, resulting in improved performance in this specific task. This also explains why Alpaca-Chinese-7B's performance on mental health support tasks was close to that of CBT-LLM.
Furthermore, while Qwen-7B possesses advantages in multilingual processing, its performance on tasks related to CBT structured interventions was slightly inferior to CBT-LLM. This might be because Qwen-7B's training data and optimization strategies lean more toward broad multilingual processing.

\begin{table}[t]
\renewcommand{\arraystretch}{1.2}
\begin{tabular}{c|ccc}
\hline
\textbf{CBT-LLM Backbone}    & \textbf{Rele.} & \textbf{Stru.} & \textbf{Help.} \\ \hline
Alpaca-Chinese-7B & 1.732             & 1.508             & 1.408             \\
\textbf{Baichuan-7B}  & \textbf{1.734}          & \textbf{1.644}          & \textbf{1.432}          \\ \hline
\end{tabular}
\caption{Human evaluation by professional raters for relevance measure (Rel.), CBT structure measure (Stru.) and helpfulness measure (Help.).}
\label{table:her}
\end{table}

\begin{figure*}[t]
    \centering
    \includegraphics[width=1\linewidth]{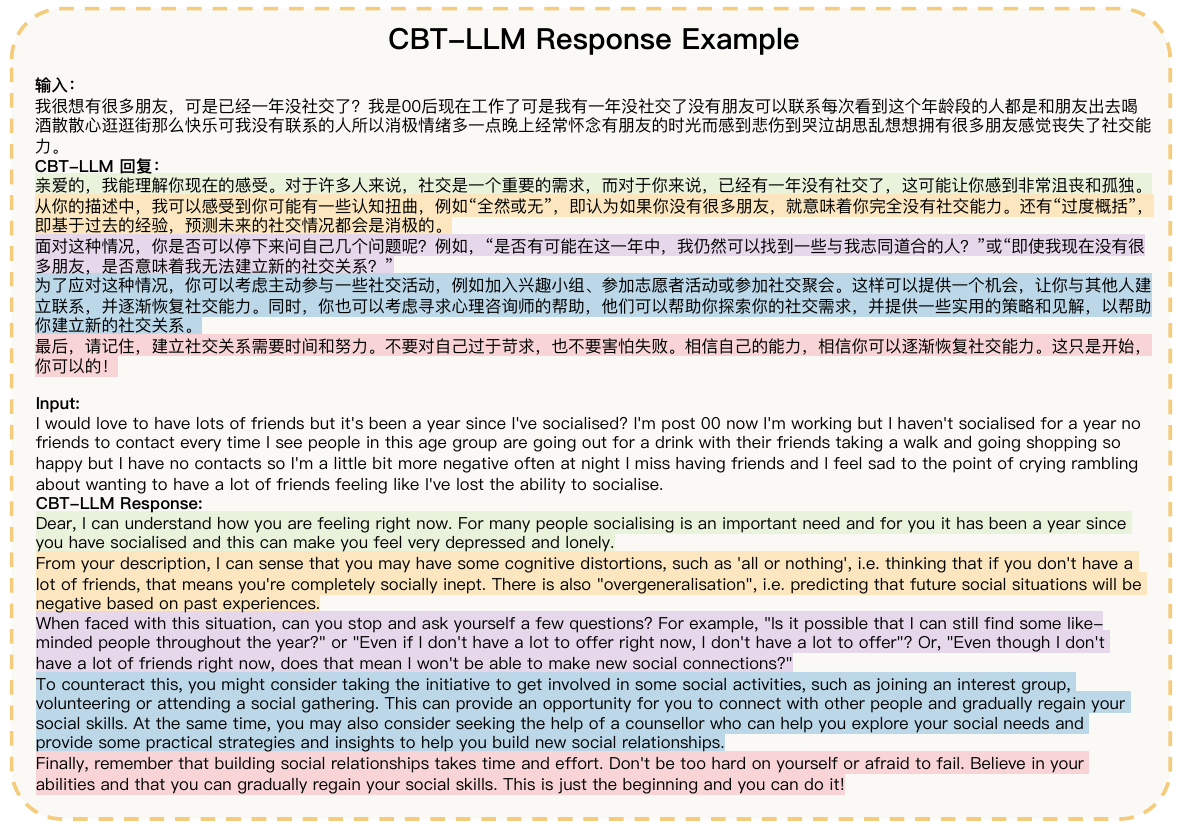}
    \caption{Five primary sections are distinctly highlighted using different colors: 1. Expression of validation and empathy (\colorbox{mygreen}{Green}), 2. Identification of Key thought or beliefs (\colorbox{myyellow}{Yellow}), 3. Introduction of challenges or reflections (\colorbox{mypurple}{Purple}), 4. Provision of strategies or insights (\colorbox{myblue}{Blue}), and 5. Encouragement and foresight (\colorbox{myred}{Red}).}
    \label{fig:case}
\end{figure*}

\subsection{Human Evaluation}
To deeply evaluate the quality of the model's responses generated based on CBT in psychotherapy counseling, we designed a comprehensive manual evaluation framework. To ensure the selection was representative, we randomly extracted 100 samples from the test set. Our evaluation team consisted of four senior psychology students and an experienced psychotherapist to ensure accuracy and professionalism. During the evaluation process, each entry consisted of three parts: a question title, a description, and an answer text. Evaluators were required to score each answer based on the following three metrics: (1) \textbf{Relevance Measure}, focusing on the association degree between the answer and the question, with scores ranging from 0-2, representing from not relevant to fully relevant; (2) \textbf{CBT Structure Measure}, assessing whether the answer adheres to specific CBT structures and principles, with scores ranging from 0-2, indicating from not adhering to structure to fully adhering; and (3) \textbf{helpfulness Measure}, assessing the applicability and usefulness of the answer from a psychotherapy perspective, with scores ranging from 0-2, representing from low applicability to highly applicable. These three metrics aim to comprehensively and deeply assess the model's application performance in the field of psychotherapy counseling. The results of the automatic evaluation are shown in the table \ref{table:her}.

The findings shown in Table \ref{table:her} indicate that Baichuan-7B marginally outperforms Alpaca-Chinese-7B in all aspects, particularly in adhering to CBT frameworks and providing helpful responses in a psychotherapeutic context.
% From Table \ref{table:her}, Baichuan-7B marginally surpasses Alpaca-Chinese-7B across all metrics. Specifically, while both models demonstrate comparable relevance, Baichuan-7B exhibits superior adherence to CBT structures by a notable margin. Additionally, in terms of helpfulness in psychotherapy context, Baichuan-7B again holds a slight edge, reinforcing its robustness in psychotherapeutic response generation.

\subsection{Case Study}
As depicted in Figure \ref{fig:case}, the CBT-LLM case study delved into addressing a user's query concerning social isolation and loneliness. The CBT-LLM model, grounded in cognitive-behavioral therapy principles, adopts a multidimensional approach. It begins by acknowledging and empathizing with the user's emotions and offering emotional support. Subsequently, the model identifies potential cognitive biases, such as overgeneralization and all-or-nothing thinking, and provides practical strategies, emphasizing the significance of social engagement and recommending professional psychological counseling. In conclusion, the model offers positive encouragement, emphasizing the individual's potential to overcome challenges. This case underscores the unique strengths of the CBT-LLM model in comprehending, guiding, and motivating users, further validating its effectiveness in delivering psychological advice and support.

\section{Conclusions and Future Work}
In this study, we presented a pioneering approach in the realm of psychological health support, bridging the gap between LLM and CBT. By introducing a CBT-specific prompt and crafting the tailored CBT QA dataset for the Chinese mental health landscape, we were able to fine-tune a large language model, thereby establishing the CBT-LLM. Empirical analyses and evaluations reaffirmed the robustness of our model, with it excelling in generating structured, professional, and highly relevant responses for psychological health support tasks.

In the future, we will explore two directions. First, beyond CBT, integrating methodologies from therapies like ACT and DBT can create a more comprehensive model, catering to diverse therapeutic needs. Secondly, transitioning from single-turn Q\&A to multi-turn dialogues will better mimic real-world counseling sessions, enhancing the realism and depth of model-patient interactions.

\section{Limitations}
The methodology of this study, while innovative, does not incorporate well-defined annotations for cognitive distortions, relying entirely on the generative capabilities of the model. This approach could lead to the generation of inaccurate types of cognitive distortions, potentially diminishing the usefulness and relevance of the model's responses. The absence of a guided annotation framework to accurately identify and categorize cognitive distortions might not only affect the precision of the advice but also the overall effectiveness of the CBT process facilitated by the model.

Furthermore, the attempt to encapsulate the comprehensive process of CBT into a single response, although aimed at efficiency, may inadvertently create a sense of pressure for the users. This is particularly true in scenarios where the model generates consecutive questions as part of the therapy process. The compactness of delivering the CBT process in one go could overwhelm users, detracting from the experience. 

\section*{Ethical Statement}
% 适当再补充一些内容
In line with the data copyright protocols delineated by the PsyQA \cite{sun-etal-2021-psyqa}, we will publicly release the CBT QA dataset for research purposes only. All questions from online mental health forums have been anonymized to protect participant privacy. Furthermore, researchers should clarify that the questions in the dataset originate from online mental health forum, and the responses are generated by ChatGPT, not professionals. Therefore, this work cannot provide any therapeutic recommendations or diagnostic statements.

\section*{Acknowledgements}
We are grateful to the psychotherapists and psychology students for human evaluation. We thank Yanyan Jiang and Qingning Lin for their suggestions and feedback. We also thank \citet{sun-etal-2021-psyqa} and \citet{llama-factory} for the public access of their datasets and efficient fine-tuning tools.

\section*{Bibliographical References}
\bibliographystyle{lrec-coling2024-natbib}
\bibliography{lrec-coling2024-example}

% 附录
% \section{Language Resource References}
% \bibliographystylelanguageresource{lrec-coling2024-natbib}
% \bibliographylanguageresource{languageresource}
\appendix
\section{Human Evaluation Guidelines}

This appendix provides the guidelines for the human evaluation of model-generated responses in the context of Cognitive Behavioral Therapy (CBT) based psychological counseling. Each evaluated entry should consist of three parts: the question, its description, and the answer text. Evaluators are required to assess each answer according to the following three metrics.

\begin{itemize}
    \item \textbf{Relevancy Metric}: This metric evaluates the connection degree between the model answer and the posed question, focusing on whether the answer specifically addresses the question's theme or core points, rather than providing unrelated or off-topic responses. The evaluation is based on a comparison of the question's keywords, themes, and the content of the model's answer. The scoring is as follows: 0 for irrelevant, 1 for partially relevant, and 2 for fully relevant answers.
    
    \item \textbf{CBT Structure Metric}: This metric assesses if the model's answer adheres to the specific structure and principles of Cognitive Behavioral Therapy (CBT), which involves identifying and challenging unhelpful thought patterns, offering alternative or more beneficial ways of thinking, and possibly providing behavioral advice. The evaluation focuses on whether the answer reflects such a structure, with scores assigned as 0 for answers not adhering to the structure, 1 for partially adhering, and 2 for fully adhering.
    
    \item \textbf{Beneficial Metric}: This metric evaluates the answer's applicability and benefit from a psychological counseling perspective. Not all technically correct answers are beneficial in a psychological counseling context. This assessment determines whether the answer provides genuine help, support, or guidance, while avoiding potentially harmful, misleading, or confusing information. The scoring is as follows: 0 for low applicability, 1 for some applicability, and 2 for high applicability.
\end{itemize}

\end{document}